\documentclass[journal]{IEEEtran}

\usepackage{amsmath}
\usepackage{graphicx}
\usepackage{amssymb}
\usepackage{booktabs}
\usepackage{algorithm}  
\usepackage{algorithmic}

\usepackage{multirow}
\usepackage{tabularx}

\graphicspath{{figs/}}

\begin{document}
	
		\title{Parallel Medical Imaging for Intelligent Medical Image Analysis: Concepts, Methods, and Applications}
	\author{Chao~Gou*,~\IEEEmembership{Member,~IEEE,}~Tianyu~Shen*,~Wenbo~Zheng,~Huadan~Xue,~Hui~Yu,~Qiang~Ji,~\IEEEmembership{Fellow,~IEEE,}~Zhengyu~Jin,~and~Fei-Yue~Wang,~\IEEEmembership{Fellow,~IEEE}
		
		\thanks{This work was supported in part by the National Natural Science Foundation of China under Grant 61806198.
		}
		\thanks{* Equal contribution. C. Gou (Corresponding Author) is with School of Intelligent Systems Engineering, Sun Yat-sen University, Guangzhou 510275, China.
			T. Shen, W. Zheng and F.-Y. Wang are with the State Key Laboratory for Management and Control of Complex Systems, Institute of Automation, Chinese Academy of Sciences, Beijing 100190, China. T. Shen is also with University of Chinese Academy of Sciences, Beijing 100049, China. W. Zheng is also with School of Software Engineering, Xi'an Jiaotong University, Xi'an 710049, China. 
			H. Xue and Z. Jin are with Department of Radiology, Peking Union Medical College Hospital, Chinese Academy of Medical Sciences and Peking Union Medical College, Beijing 100730, China.
			H. Yu is with School of Creative Technologies, University of Portsmouth, Portsmouth, PO1 2DJ. Q. Ji is with Department of Electrical, Computer, and Systems Engineering, Rensselaer Polytechnic Institute. F.-Y. Wang is also with Qingdao Academy of Intelligent Industries, Qingdao 266000, China. (e-mail: gouchao@mail.sysu.edu.cn, shentianyu2016@ia.ac.cn, zwb2017@stu.xjtu.edu.cn, bjdanna95@163.com, hui.yu@port.ac.uk, qji@ecse.rpi.edu, jin\_zhengyu@163.com, feiyue.wang@ia.ac.cn)
		}
		
	}

	\markboth{IEEE/CAA JOURNAL OF AUTOMATICA SINICA,~Vol.~X, No.~X, X~X}%
	{Gou \MakeLowercase{\textit{et al.}}: Parallel Medical Imaging}
	
	\maketitle
	
	\begin{abstract}
		Data-driven artificial intelligence technologies have made much progress in medical image analysis in the last decades.
		However, it still remains challenging due to its distinctive complexity of acquiring and annotating image data, extracting medical domain knowledge, and explaining the diagnostic decision for medical image analysis. In this paper, we propose a data-knowledge-driven framework termed as Parallel Medical Imaging (PMI) for intelligent medical image analysis based on the methodology of interactive ACP-based parallel intelligence. In the PMI framework, computational experiments with predictive learning in a data-driven way are conducted to extract medical knowledge for diagnostic decision support. Artificial imaging systems are introduced to select and prescriptively generate medical image data in a knowledge-driven way to utilize medical domain knowledge. Through the closed-loop optimization based on parallel execution, our proposed PMI framework can boost the generalization ability and alleviate the limitation of medical interpretation for diagnostic decisions. Furthermore, we illustrate the preliminary implementation of PMI method through the case studies of mammogram analysis and skin lesion image analysis. Experimental results on several public medical image datasets demonstrate the effectiveness of proposed PMI.
		
	\end{abstract}
	
	
	\IEEEpeerreviewmaketitle
	
	\section{Introduction}
	\IEEEPARstart{M}{edical} image analysis aims at extracting clinically useful information from computed
	tomography (CT), positron emission tomography (PET),  magnetic resonance (MR), ultrasound, X-ray, and other modalities of images with the assistance of computers for diagnostic decision support \cite{shen2017deepSurvey,hu2018deepDetectionMedSurvey}. With urgent requirements of medical imaging, medical societies have entered a new era that medical equipments, image data, domain knowledge, and humans including physicians and patients are coupled in the large scale cyber-physical-social spaces (CPSS) \cite{wang2013intelligent,wang2020Parallel}. Hence, vision-based medical image analysis is becoming an increasingly prominent role at many clinical workflow stages from screening and diagnosis to treatment delivery, especially in the domain of remote medical consultation. Recently, vision-based medical image analysis has achieved promising results for skin cancer diagnosis \cite{haenssle2018man,esteva2017skin_nature,wang2019parallelSkin,zheng2019relationSkin,zhang2019attention}, red lesion detection in fundus images \cite{orlando2018fundusRedLesion}, mammography analysis \cite{bejnordi2017diagnostic,antari2018fully,shen2020simultaneous,shen2020Hierarchical} and pulmonary nodule detection \cite{zhu2018deeplung,winkels2019pulmonary}. However, there are still challenges for vision-based medical image analysis. Firstly, these data-driven techniques require a large scale of effective medical images annotated by domain experts or radiologists. Secondly, conventional methods for medical imaging are built in a data-to-knowledge way where algorithms are learned from existing training samples in a bottom-up manner without any feedback or interaction to utilize the medical domain knowledge. Last but not the least, there is a limitation in interpretability for the final medical diagnostic decisions made by learned "black-box" models especially for non-linear deep neural networks.
	
	ACP methodology was first proposed in \cite{wang2004parallel} for modeling, managing and controlling the complex systems. It consists of \emph{Artificial societies}, \emph{Computational experiments} and \emph{Parallel execution}. The ACP-based parallel intelligence is one form of intelligence generated from the interactions and executions between physical and artificial systems \cite{wang2016steps}. As part of parallel intelligence, the parallel learning framework was presented in \cite{li2017parallel} to address issues of data collection and policy exploring in the current machine learning framework. Parallel learning combines descriptive learning, predictive learning, and prescriptive learning into a uniform evolutionary framework to optimize the learning system by self-boosting \cite{li2018crossroad}.
	
	Inspired by the interactive ACP-based parallel intelligence, we propose a data-knowledge-driven framework termed as Parallel Medical Imaging (PMI) to address the aforementioned challenges in medical image analysis. 
	Firstly, as pointed by Wang \emph{et al.} in \cite{wang2016data} that evaluations of the objectives can be performed only based on data collected from physical world and virtual world, we propose to conduct computational experiments to predictively extract medical knowledge for diagnostic decision support that are explainable to humans in a data-driven way. Different from conventional medical image analysis frameworks that solely perform data-to-knowledge extraction, we further introduce artificial imaging systems to select and generate specific medical image data for data collection in a knowledge-driven way. Specifically, interactive parallel learning with a descriptive and prescriptive scheme based on the explainable knowledge is performed to achieve knowledge-to-data generation in a top-down manner that allows for boosting the performance of decision model. In addition, the data-knowledge-driven parallel evolution can enable effective large scale data collection and enhance the interpretability of diagnosis.
	
	

	\section{Preliminary}\label{ACP-paralele}
	\subsection{ACP methodology}
	The ACP methodology was initially proposed by Wang \emph{et al.} \cite{wang2004parallel} for effective modeling and controlling of the complex systems. It consists of Artificial societies (A), Computational experiments (C) and Parallel execution (P). The key idea of ACP is to combine Artificial societies, Computational experiments, and Parallel execution to turn the virtual artificial space into another space for solving complex problems \cite{wang2016AlphaGo}. It is further extended to ACP-based parallel intelligence that is defined as one form of intelligence generated from the interactions and executions between physical and artificial systems \cite{wang2016steps}. ACP-based parallel intelligence is becoming an increasingly important research topic and is widely applied in various circumstances such as social computing, traffic management and control,  ethylene production management, and autonomous driving \cite{wang2016steps,wang2010parallel,wang2013intelligent,wang2017parallelDriving}.
	
	\subsection{Parallel vision framework}
	ACP methodology is further extended to the computer vision community as a Parallel Vision (PV) framework for better perceiving and understanding complex scenes in \cite{wang2017parallelVision}. Inspired by ACP methodology, PV contains three parts including artificial systems, computational experiments and parallel execution. From the perspective of PV, the vision models are on-line optimized through parallel execution with a virtual/real interactive policy. From the perspective of PV, existing work of learning-by-synthesis \cite{pezeshk2017seamless,frid2018gan,gou2017joint} is part of PV with respect to artificial systems and computational experiments. Parallel execution aims to construct a closed loop driven by a large scale of "big data" to boost the performance of vision systems. As a result, on-line learning through parallel execution allows the perception model to be continuously optimized.
	
	\subsection{Parallel learning} \label{paraLearning}
	By taking data, knowledge, and action into a closed loop, a parallel learning framework was introduced in \cite{li2017parallel} to alleviate the limitation in data collecting and policy exploring of existing machine learning frameworks. Based on ACP methodology, the parallel learning framework can capture the mutual dependency between data and action in an artificial system parallel to the physical system from observations. Parallel learning combines descriptive learning, predictive learning and prescriptive learning to effectively collect/generate data and guide the implementation of complex learning systems \cite{li2018crossroad}. 
	
	In particular, predictive learning urges the decision system to extract informative knowledge based on collected data. For descriptive learning, it forms a self-consistent artificial system to generate new labeled data following the distribution of observed data in real system with minimum human intervention. Moreover, prescriptive learning allows to guide the system to collect specific data in a supervised manner with descriptive or prior knowledge \cite{ullrich2008descriptive}. Refer to \cite{li2017parallel,li2018crossroad} for more details.
	
	\section{Parallel medical imaging}\label{PMI_sec}
	Conventional medical image analysis frameworks extract clinical knowledge from image data in a bottom-up manner where the model learning is driven by data ignoring the prior medical knowledge. However, in the field of medical imaging, domain knowledge plays a critical role for data collection and diagnosis decision support. Properly utilizing medical knowledge in a top-down manner can not only improve the diagnosis but also enhance the interpretability of diagnostic decision. Inspired by the ACP-based parallel intelligence frameworks and systems, we propose a data-knowledge-driven framework termed as parallel medical imaging (PMI) for medical image analysis. 
	
\begin{figure*}[!tbp]
		\centering
		\includegraphics[width = 6.2in]{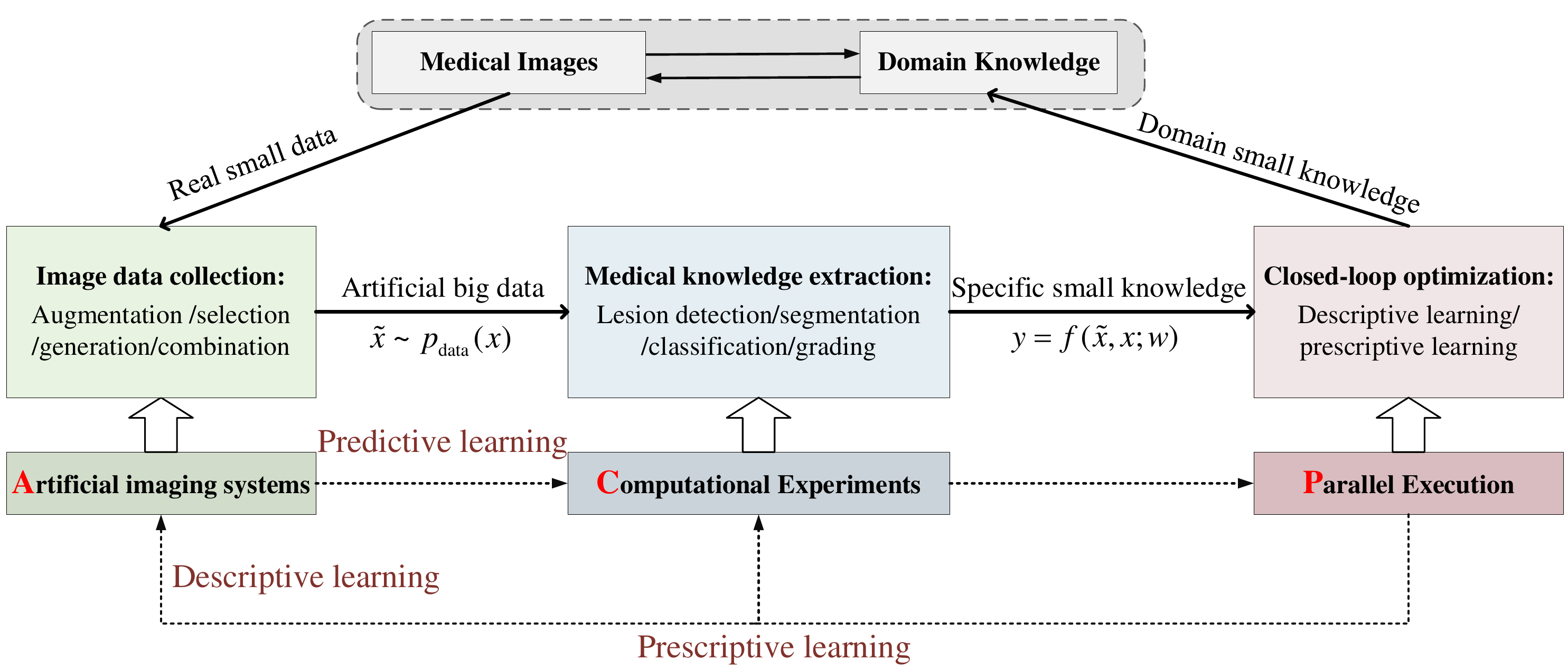}
		\caption{Overall framework of PMI.}
		\label{PMIFramework}
	\end{figure*}

The overall framework of proposed PMI is shown in Fig. \ref{PMIFramework}. 
	Two major parts of medical image and mdomain knowledge are coupled in PMI by ACP methodology and parallel learning. 
	The key point is to select and generate image data which are representative to extract desired medical knowledge for final diagnostic decision. 
	
	Firstly in the stage of artificial systems, raw images are collected, followed by variation operators such as augmentation, selection and reproduction with generation for large scale of image data collection. In this work, inspired by the key idea of evolutionary optimization through the interactions and executions between physical and artificial systems, we introduce artificial imaging systems (AIS) parallel to physical ones. Particularly, a self optimizing AIS can be constructed through descriptive learning. According to relevant knoeledge and the distribution ($ p_{\text{data}}(x) $) of acquired real small data, abundant artificial data ($ \tilde{x}\sim p_{\text{data}}(x) $) are generated. 
	Secondly, computational experiments with predictive learning are conducted for data-to-knowledge extraction: $ y=f(\tilde{x}, x; w) $. $ (\tilde{x}, x) $ is the combinbation of real small data and artificial big data in AIS, and $ y=f(\cdot; w) $ is the mapping represented by the visual models in the computational experiments.
	Finally, in the stage of parallel execution, prescriptive learning is adopted to guide the data generation in AIS based on the predictively extracted or prior medical knowledge where knowledge-to-data is achieved. This step can also enhance the interpretability of decision. In addition, descriptive learning is adopted in AIS to guide the data selection and generation based on the captured data distribution and knowledge. As a result, final effective diagnosis and prognosis can be achieved through extracted knowledge with enhanced interpretability. Hence, PMI can jointly employ the image data distribution and medical knowledge through bottom-up and top-down learning and inference for final clinical decision. It can reduce the dependence on annotated images and alleviate the limitation of medical interpretation for diagnostic decision. More details are given in the following subsections.

	\subsection{Image data collection}
	
	For medical imaging, large scale of image data with accurate annotations is critical for the performance of learning-based methods. Parallel imaging framework was introduced in \cite{wang2017parallelImaging} for image generation for PV \cite{wang2017parallelVision} to tackle the problems of complex vision systems. However, compared with natural image analysis, medical image analysis requires a higher level of expertise for interpretation and labeling. In addition, it is not easy to collect image data from medical institutions or imaging communities since they should be in accordance with the specific security and privacy policies. Moreover, some lesion types and abnormalities have a very low rate of occurrence in the general population \cite{pezeshk2017seamless,shen2019learning}. It is thus more time-consuming and costly to collect effective training data which makes medical imaging remain a challenging task. 
	
	In this work, as shown in Fig. \ref{PMIFramework}, from the data perspective of parallel intelligence \cite{wang2016steps,wang2017parallelImaging}, real medical images with annotations are kind of 'small data'. Through effective reproduction and variation operation such as conventional augmentation, active selection, and generation by introduced artificial imaging systems, a set of 'big data' with real and synthetic images is formed for conducting computational experiments for medical knowledge extraction. 
	
	\subsubsection{Augmentation and selection of real images}
	In the step of image data collection, small and/or imbalanced real images for training can be augmented. Similar to conventional methods, rotation, scaling, flipping, translation and noise addition can be applied for medical image augmentation \cite{esteva2017skin_nature,frid2018gan,mikolajczyk2018data,zheng2019relationSkin}. 
	
	The performance of learning-based methods for medical image analysis depends not only  on the size but also the representativeness of labeled images. However, due to a lack of standardization in imaging and acquisition for medical images, selecting representative training samples for computational experiments remains a challenging task. In this framework, suitable selection of real images is performed to address this challenge. To this end, simple unsupervised/semi-supervised can be applied for data selection. In addition, active learning that aims at using limited medical images for disease classification can be developed. Active learning iteratively selects the most informative samples through the interaction between experts and computer. In active learning, the key is to develop a criterion for uncertainty in the sample selection process. In \cite{zhou2017fine_active}, \emph{entropy} and \emph{diversity} are adopted to indicate the power of candidate patches in elevating the performance of the current CNN model. For the $i$-th  patch of $k$-th candidate denoted by $x_{k}^{i}$, the prediction is $p_{k}^{i}$ and its entropy is formulated as below:
	\begin{equation}\label{entropy}
	e(x_{k}^{i})=- \sum_{c=1}^{N}p_{k}^{i,c}\log p_{k}^{i,c},
	\end{equation}
	where $c=1,2,..., N$ is the possible class. The \emph{diversity} between sample patch $x_{k}^{i}$ and $x_{k}^{j}$ is defined as below:
	\begin{equation}\label{diversity}
	d(x_{k}^{i},x_{k}^{j})=\sum_{c=1}^{N}(p_{k}^{i,c}-p_{k}^{j,c})\log \frac{p_{k}^{i,c}}{p_{k}^{j,c}}.
	\end{equation}
	According to \cite{zhou2017fine_active}, the sample patch with higher entropy and higher diversity are expected to be selected to elevate the model performance. 
	
	\subsubsection{Generation of synthetic images}\label{synthesis}
	\begin{figure}[!t]
		\centering
		\includegraphics[width = 3.4in]{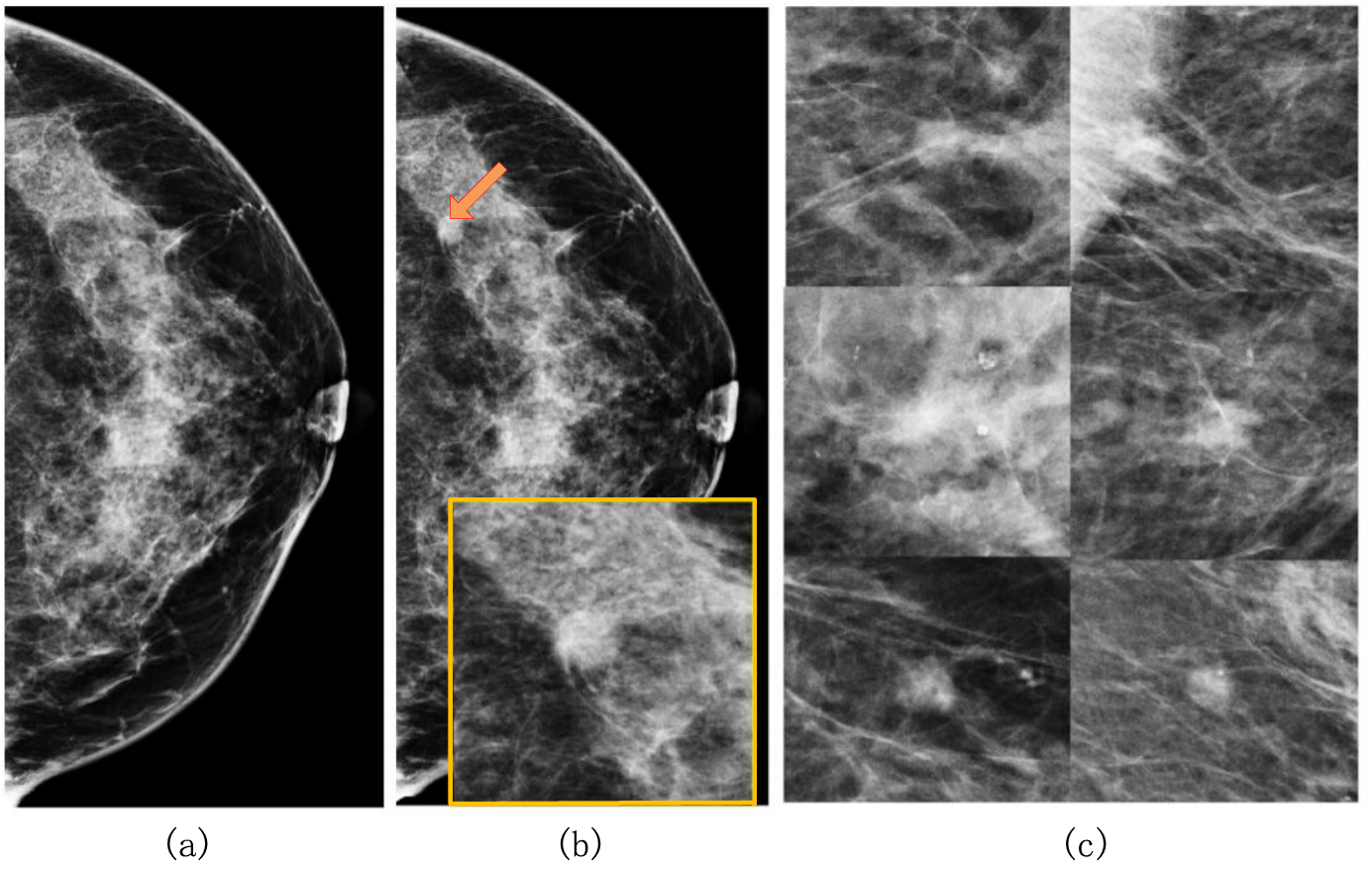}
		\caption{Synthesis by insertion. (a) Normal mammogram, (b) Insert mass, (c) Simulated lesions by diffusion limited aggregation \cite{rashidnasab2013simulation}.}
		\label{mamm_insertation}
	\end{figure}
	
	To utilize the medical domain knowledge, we propose to apply descriptive learning and design artificial imaging systems parallel to real imaging systems that can generate synthetic and specific medical images following the distribution of real ones. Many techniques for generating new synthetic medical images in our proposed framework of artificial imaging systems can be applied. They typically fall into three categories. 
	
	In the first one, new lesions are mathematically simulated based on various deformation, followed by inserting into the raw projection data or reconstructed clinical images, such as mammography \cite{rashidnasab2013simulation} and lung nodules \cite{li2009three}. An example from \cite{rashidnasab2013simulation} is illustrated in Fig.\ref{mamm_insertation}.
	To assure the realism of the characteristics of the artificial samples, real lesions can be extracted and inserted to the same or different images \cite{pezeshk2015seamless}. 
	
\begin{figure}[!t]
		\centering
		\includegraphics[width = 3.4in]{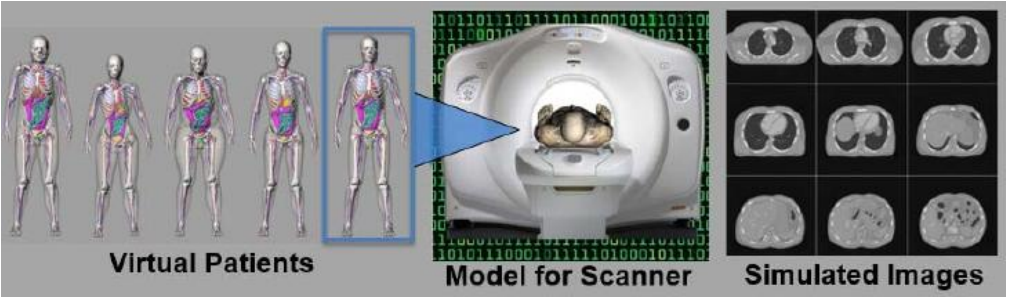}
		\caption{Computed tomography (CT) synthesis through an XCAT phantom\cite{segars2018applicationXCAT}.}
		\label{XCAT_simulation}
	\end{figure}

	In the second one, virtual images are simulated through computer graphics based on abstraction of the prior medical knowledge. Particularly, synthetic images are generated by selection of simulation parameters of models under controlled hypothetical imaging conditions \cite{frangi2018simulation}. In \cite{segars2018applicationXCAT,abadi2017modeling}, computerized phantom (eXtended CArdiac-Torso, XCAT) is served as a virtual patient, followed by feeding into an artificial imaging system with an accurate computerized model, which can generate photorealistic CT image data with patient-quality as show in Fig.\ref{XCAT_simulation}. 
	
\begin{figure}[!t]
		\centering
		\includegraphics[width = 2.9in]{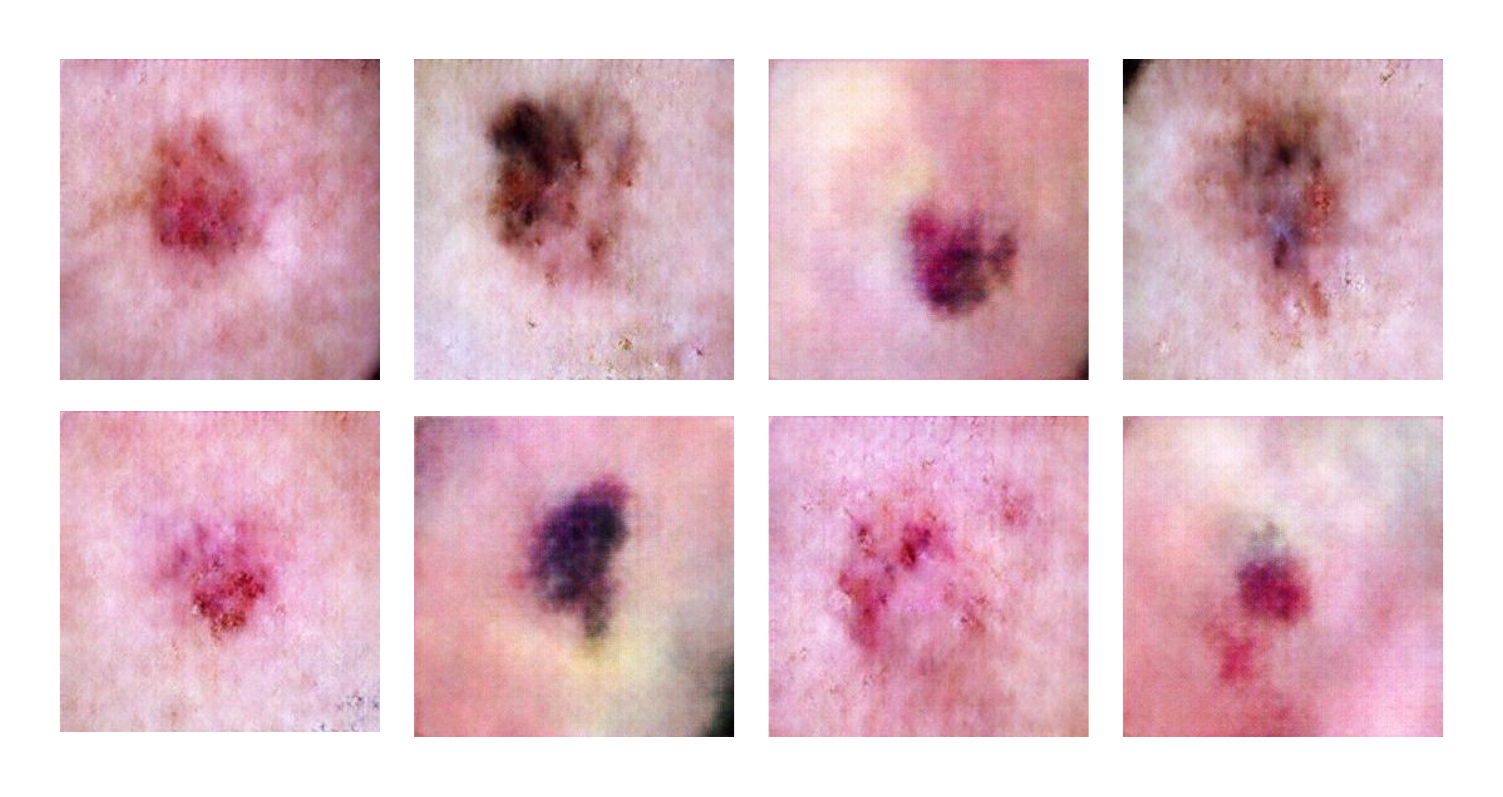}
		\caption{Synthesized images with skin lesions based on GAN}
		\label{sampleSkinSynthesis}
	\end{figure}

	In the third one, generative models for image synthesis can be learned in the artificial imaging systems. In \cite{chartsias2018multimodal}, the authors propose a model of fully convolutional neural networks for MRI synthesis. This model learns to input modalities into a shared modality-invariant latent space which allows it to benefit from additional input modalities and robust to missing data. Recently, adversarial learning for the generative model is widely used for medical image synthesis \cite{shen2019learning,bentaieb2018adversarial,wang2019parallelSkin}. Some sample synthesized skin lesions generated based on GANs \cite{wang2019parallelSkin} are shown in Fig.\ref{sampleSkinSynthesis}.
	In this work, effective generator of GANs can be utilized into the step of artificial imaging systems.
	
	\begin{figure}[!htbp]
		\centering
		\includegraphics[width = 2.5in]{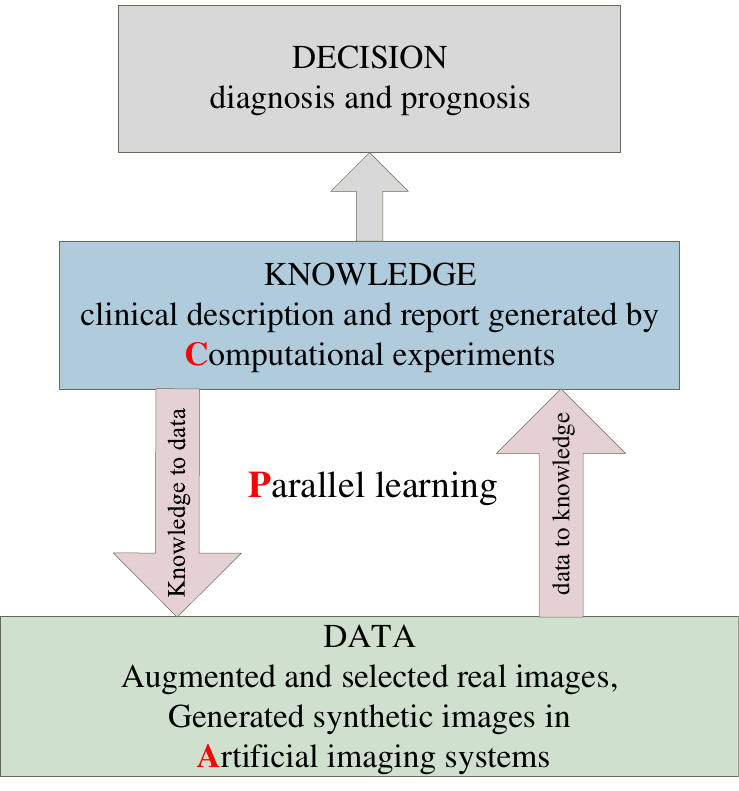}
		\caption{Data-knowledge-driven for decision pyramid in PMI.}
		\label{dataKnowledgePyramid}
	\end{figure}
	
	\subsection{Medical knowledge extraction} \label{medicalKnowledge}
	
	Conventional methods of turning data into medical knowledge rely on visual analysis and interpretation by a domain expert or radiologist in order to find useful patterns in data for decision support \cite{holzinger2014knowledge}.  As pointed in radiomics \cite{gillies2015radiomics,lambin2017radiomics}, effective conversion of images to mineable data supports the diagnostic decision. In this work, after effective image collection, computational experiments with predictive learning are conducted to extract medical knowledge in PMI. Hence, medical knowledge extraction from images  is also a part of radiomics.
	
	For this research topic in PMI, any information (e.g. mass shape, margin, density, location) about the patient's ultrasonic signs, X-ray findings and other related image-based medical descriptions are termed as 'symptom'. Computational experiments with predictive learning try to perform effective diagnosis. To achieve this goal, medical knowledge needs to be extracted by studying the relationships of obligatory proving or excluding symptoms for diagnosis in books and in practical experience. These certain information about relationships that exist between symptoms and diagnoses, symptoms and symptoms, diagnoses and diagnoses and more complex relationships of combinations of symptoms and diagnoses to a symptom or diagnosis are formalizations of what is called medical knowledge \cite{seising2003medical}. 
	
	Predictive learning was originally inspired by the cognitive psychology study that how children construct knowledge of the world by interacting with it \cite{wang2017parallelDriving}. In the step of computation experiments, we perform predictive learning for the diagnosis model from collected image data for decision support. It can be simplified as part of medical knowledge extraction from image data. Conventional data-driven machine learning techniques especially deep learning models can be learned to address knowledge extraction in PMI.  Moreover, the hand-crafted features (e.g. LBP, HOG, SIFT) designed by humans using statistical formulation
	can be applied as prior domain knowledge for approximations of visual content \cite{zheng2019relationSkin,cruz2018high}.  
	
	In general, computational experiments in PMI include detection, segmentation, classification, or relationship caption for decision support for clinical applications. The detection model extracts the knowledge of rough location and size of the lesion area. Subsequently, the segmentation model extracts the detailed shape and margin information of the lesion. Finally, the knowledge of pathological types and assessment categories are obtained through the classification task.
	
	\begin{figure*}[!t]
		\centering
		\includegraphics[width=0.9\linewidth]{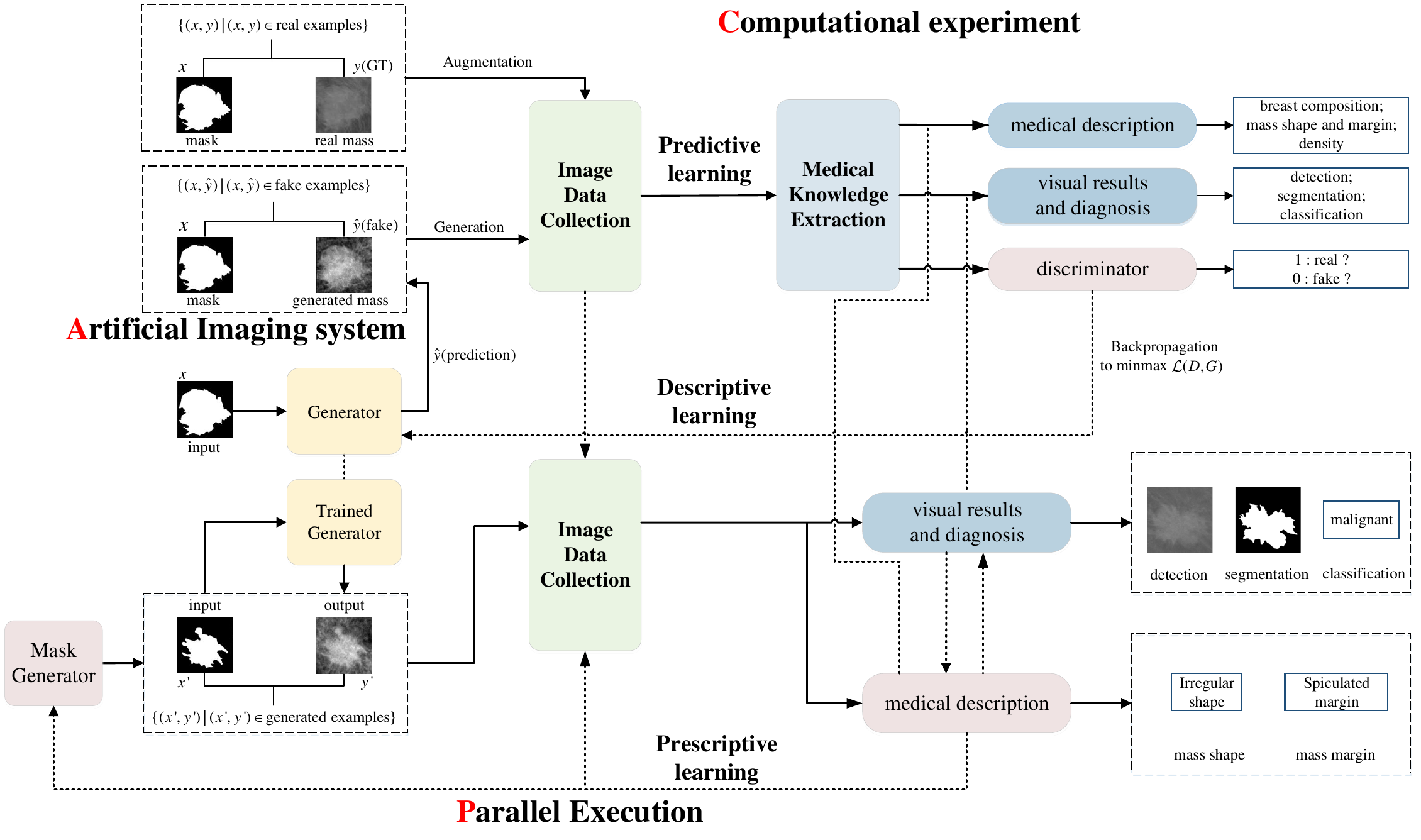}
		\caption{GANs-based PMI framework for breast mass analysis.}
		\label{parallelMedGAN}
	\end{figure*}

	\subsection{Closed-loop optimization with parallel learning}	
	As shown in Fig. \ref{dataKnowledgePyramid}, we introduce parallel learning to take advantage of bidirectional optimization between medical image data and clinical description/representation of medical knowledge. In PMI, predictive learning can achieve effective data-to-knowledge extraction through a bottom up manner.. Different from traditional diagnosis of treating medical images as pictures intended solely for visual interpretation, conversely, through a top-down inference, the extracted medical knowledge can be used for guiding the image generation as well as increasing the interpretability of future diagnosis. As described in subsection \ref{paraLearning}, we employ descriptive and prescriptive learning of parallel learning to improve the model generalization ability and enhance the interpretation for medical diagnosis decision. 
	
	\subsubsection{Descriptive learning}
	 Descriptive learning aims to devise models to explain and predict learning results \cite{ullrich2008descriptive}. In this work, it urges the introduced artificial imaging system to generate new images that follow the distribution of observed data. The descriptive learning process allows for learning features from unlabeled data in a semi-supervised or unsupervised manner.  Adversarial learning of GAN for image generation can be seen as a special case of descriptive learning where the objective is to minimize the difference of distribution for real between generated images\cite{li2017parallel,li2018crossroad}. 
	 
	 Taking mass image generation for an example \cite{shen2019learning}, we introduce adversarial learning as part of descriptive learning to implicitly learn the mass image distribution $p_{\text{mass}}$ from real image samples. GAN contains a generator $G$ and a discriminator $D$, and the output of discriminator $D$ can be descriptively used for the optimization of generator $G$. For the input $x$, $D(x)$ represents the probability of being a real mass image. For the input $z$ from a simple distribution $p_{z}$, $G(z)$ represents the generated synthetic image. The basic loss function for adversarial learning is a two-player minimax game formulated as below:
	 \begin{equation}\label{obj}
	 \begin{split}
	 \displaystyle\mathop{\mathrm{min}}\limits_{G}
	 \displaystyle\mathop{\mathrm{max}}\limits_{D} \{f(D,G)=\,&{\mathbb{E}}_{x\sim p_{\text{mass}}(x)}[\log D(x)] +\\&{\mathbb{E}}_{z\sim
	 	p_{z}(z)}[\log (1-D(G(z)))]\},
	 \end{split}
	 \end{equation}
	 where $\mathbb{E}$ represents the expectation. We can rewrite this loss function as below:
	 \begin{equation}
	 f_{S}=\, {\mathbb{E}}[\log P(S=\text{real}|X_{\text{real}})] +{\mathbb{E}}[\log P(S=\text{fake}|X_{\text{fake}})],
	 \end{equation}
	 where $P(S|X)=D(X)$ and $X_{\text{fake}}=G(z)$. And $D$ is trained to maximize $f_{S}$ which denotes it assigns to the correct mass image source. $G$ is trained to minimize the second term of $f_{S}$. It is worth noticing that the key idea of descriptive learning is to model the medical image distribution, perception and reasoning based on the observation in real world. Hence, the descriptively learned generator through adversarial learning can be generalized to artificial imaging systems.
	
	\subsubsection{Prescriptive learning}
	According to the definition in \cite{ullrich2008descriptive,li2017parallel,li2018crossroad}, prescriptive learning is concerned with guidelines that describe what to do in order to generate specific outcomes. They are often based on descriptive theories or derived from prior knowledge. In this work, we achieve knowledge-to-data generation and enhance interpretability through prescriptive learning of parallel learning. According to the ACP methodology, we perform parallel execution with prescriptive learning to guide the artificial medical imaging systems to collect specific representative image data based on the extracted or prior medical descriptions and knowledge. 
		
	For instance, based on the prior medical knowledge that mammograms with spiculated and irregular mass are mostly malignant, we can prescriptively generate various irregular and spiculated mass images with associated pleomorphic calcifications for malignant breast cancer analysis in mammograms \cite{kim2018icadx}. As a result, visual interpretation on the diagnostic results is enhanced through prescriptive learning which effectively capture the relationship between malignancy and interpretability.

	\section{Case Studies of PMI}\label{GAN_implementation}
	\subsection{Analysis of mass in mammograms}
	To validate the effectiveness of the proposed PMI framework, we further perform a case study of mammogram analysis in this subsection. The  clinical descriptive details from standard Breast Imaging Reporting and Data System (BI-RADS) \cite{d2013acrBIRADS} are illustrated in Table.\ref{table1} and Table.\ref{table2} that explicitly inform the domain knowledge description. Similar to the work in \cite{kim2018icadx}, after capturing the relationship between the malignancy and clinical description as listed in Table \ref{table2}, diagnosis with interpretability can be enhanced. For visual results and diagnosis, the visual diagnosis models is trained for visual information extraction like detection, segmentation and classification.  Built upon PMI, we perform an implementation based on GANs with image data collection, medical description of knowledge and parallel learning. The overall framework is illustrated in Fig. \ref{parallelMedGAN}. Due to page limitation, we only study the problem of local X-ray breast mass classification (benign/malignant) for diagnosis. Another case study of X-ray breast mass segmentation based on ACP methodology can be referred to \cite{shen2019learning}.
	
	\begin{table*}[!tbp]
		\renewcommand\arraystretch{1.1}
		\centering
		\caption{Clinical description for mammography: breast composition, mass shape and margin, density.}
		\label{table1}
		\begin{tabular}{p{2.8cm}<{\centering} p{2cm} p{9cm}}
			\hline		
			\multirow{4}*{Breast composition}
			&\multicolumn{2}{l}{a. The breast are almost entirely fatty;}     \\ 
			&\multicolumn{2}{l}{b. There are scattered areas of fibroglandular density;}     \\          
			&\multicolumn{2}{l}{c. The breasts are heterogeneously dense, which may obscure small masses;}     \\  
			&\multicolumn{2}{l}{d. The breasts are extremely dense, which lowers the sensitivity of mammography.}     \\  
			\hline 
			\multirow{3}*{masses}		
			&shape    &Oval; Round; Irregular.  \\
			&Margin   &Circumscribed; Obscured; Microlobulated; Indistinct; Spiculated.  \\
			&Density  &High density; Equal density; Low density; Fat-containing.  \\
			\hline
		\end{tabular}
	\end{table*}
	
	\begin{table}[!tbp]
		\renewcommand\arraystretch{1.1}
		\centering
		\caption{Breast Imaging Reporting and Data System (BI-RADS) Assessment Categories}
		\label{table2}
		\begin{tabular}{p{1cm}<{\centering} p{5cm}<{\centering}}
			\hline		
			Category  &Description	 \\
			\hline 
			0         &Needs additional imaging evaluation and/or prior mammograms for comparison. \\
			1         &Negative. \\
			2         &Benign finding(s). \\
			3         &Probably benign finding(s). Short-interval follow-up is suggested. \\
			4         &Suspicious anomaly. Biopsy should be considered. \\	
			5         &Highly suggestive of malignancy. Appropriate action should be taken. \\
			6         &Biopsy proven malignancy. \\
			\hline
		\end{tabular}
	\end{table}
	
	\subsection{Implementation}
\begin{figure*}[!tbp]
		\centering
		\includegraphics[width=0.9\linewidth]{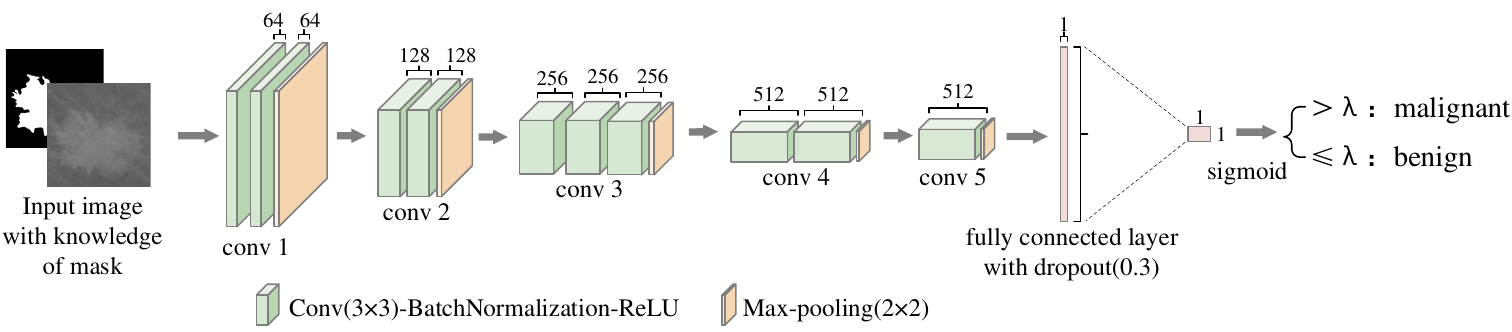}
		\caption{CNN architecture for the mass classification.}
		\label{CNN}
	\end{figure*}

	Firstly, we achieve malignancy extraction by predictive learning. In particular, we apply a simple CNN to perform predictive learning to classify the mass image with corresponding masks as malignant or benign in the step of computational experiments. The CNN architecture with details is exhibited in Fig. \ref{CNN}. Cross Entropy is used as a loss function which is computed by:
	
	\begin{equation}\label{bin_objective}
	L = -\frac{1}{n}\sum_{i=1_{(x,y)}}^{n}[l^{i}\log(p^{i})+(1-l^{i})\log(1-p^{i})],
	\end{equation}
	where $(x,y)$ denotes the pair of input, $n$ is the number of training samples, $l$ represents actual label with $1$ denoting malignancy and $0$ for the benign, $p$ is the predicted value.
	
	
	For the step of descriptive learning, inspired by the idea that adversarial learning is a special case of parallel learning \cite{li2017parallel}, we introduce a generative adversarial network structure for descriptive mass images generation in the  artificial imaging system. Specifically, a conditional GAN (cGAN) structure  is introduced for generation from given binary masks $x$ which already incorporate the shape and margin descriptive information. The generator $G$ and discriminator $D$ of GAN are trained for learning the distribution of mass images as well as a mapping $G:\{x,z\}\rightarrow y $ between masks $x$, random noise $z$, and real mass images  $y$. Inspired by \cite{isola2017image}, an U-net structure is also introduced as the generator and a PatchGAN architecture is introduced as the discriminator. To learn an effective generator $G$ in artificial imaging systems based on adversarial learning, we set the objective function as below:
	\begin{equation}\label{G_objective}
	\begin{split}
	G^{*} = arg \displaystyle\mathop{\mathrm{min}}\limits_{G}\displaystyle\mathop{\mathrm{max}}\limits_{D}&
	\,{\mathbb{E}}_{x,y}[\log D(x,y)] +\\
	&{\mathbb{E}}_{x,z}[\log (1-D(x,G(x,z)))]+\\
	&{\mathbb{E}}_{x,y,z}[||y-G(x,z)||_{1}].
	\end{split}
	\end{equation}
	And the discriminator $D$ is learned as described in \cite{goodfellow2014generative} that aims to distinguish the input as real or synthetic. $G$ and $D$ are alternatively optimized until convergence. Then we can acquire the effective generator which is part of the artificial imaging system performing synthetic data generation. The combination of the 'small' set of real images and 'big' set of synthetic ones forms a large scale of training samples that can be further fed into predictive learning for 'small' knowledge extraction.
	
\begin{figure}[!tbp]
		\centering
		\includegraphics[width=3.4in]{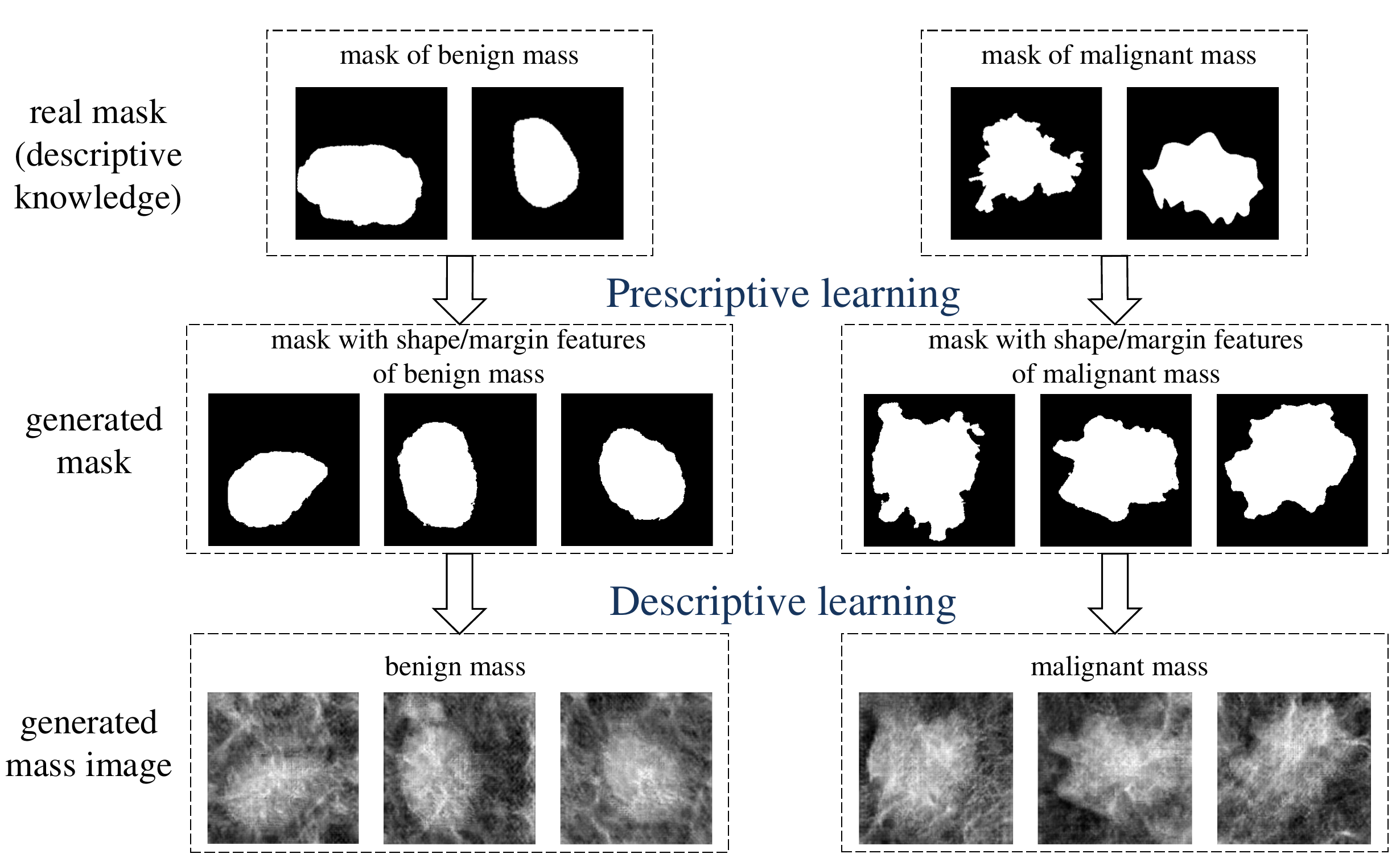}
		\caption{Qualitative results for generating binary masks and corresponding mass images by our proposed framework.}
		\label{genresult}
	\end{figure}

	As shown in Fig. \ref{dataKnowledgePyramid}, the workflow of data-to-knowledge is a bottom-up manner with the medical visual and descriptive knowledge learning from existing training samples. Conversely, through the knowledge-to-data inference in a top-down manner, the medical knowledge is used for guiding the image augmentation, selection and generation, and enhancing the interpretability of diagnosis. In this case, prescriptive learning is adopted to generate specific malignant/benign mask images with the corresponding shape and margin based on the descriptively extracted knowledge or prior medical knowledge. To this end, a deep convolutional generative adversarial network (DCGAN) \cite{radford2015DCGAN} is implemented to generate the specific binary mask. To utilize the medical knowledge that malignant mass is also with irregular shape and spiculated margin, and benign mass with oval shape and circumscribed margin \cite{kim2018icadx,d2013acrBIRADS}, we train two generators separately through adversarial learning in the prescriptive scheme for benign and malignant binary mask generation. In this work, 37 benign and 75 malignant masks are augmented into 296 benign and 600 malignant masks for training the DCGAN model. 262 benign and 409 malignant masks are obtained and used to generate the corresponding mass images through the previously trained cGAN model in the step of descriptive learning. Some generated masks are shown in Fig. \ref{genresult}. Hence, a generative models from introduced DCGAN can achieve knowledge-driven data generation. 
	
	By feeding the generated benign and/or malignant binary masks into artificial imaging systems through descriptive learning, more specific realistic-looking lesion images from interpreting conditions such as margin and shape of masses can be collected. Then we can extract more suitable medical knowledge through predictive learning in a data-driven way for final diagnosis. The overall framework jointly employs the image data collection and medical knowledge extraction in a closed loop through data-to-knowledge predictive learning and knowledge-to-data prescriptive learning. Parallel data-knowledge-driven optimization is achieved.
	
	\subsubsection{Dataset of Mammograms and Evaluation Criteria} \label{Criteria}
	Experiments are conducted in the public available dataset of INbreast \cite{INbreast} which is one of most widely used for mammogram analysis. The INbreast dataset is created by the Breast Research Group, INESC Porto, Portugal, and consists a total of 115 cases (410 images) including 107 images of cancer and 236 images of normal breast. In this work, local ROI of 107 mass images with cancers are cropped into $ 256\times 256 $ pixels along with the corresponding mask applying the same operation. A set of total 112 squared mass images is obtained because some of these cases have more than one mass and they are annotated (benign or malignant) according to the Breast Imaging Reporting and Data System (BI-RADS), which is a standard criteria developed by the American College of Radiology (ACR)\cite{INbreast} as listed in Table \ref{table2}. In this work, 36 masses with BI-RADS Category $\in \{2,3\}$ are categorized as benign, and 76 masses with BI-RADS Category $\in \{4,5,6\}$ are categorized as malignant.
	
	The performance is analyzed by measurement metrices in the binary classification problem, including overall accuracy, sensitivity,
	specificity, which are defined as 
	\begin{equation}
	\begin{split}
	&\text{accuracy (Acc.)} = \dfrac{\text{TP}+\text{TN}}{\text{TP}+\text{FN}+\text{TN}+\text{FP}},\\ 
	&\text{specificity} = \dfrac{\text{TN}}{\text{TN}+\text{FP}},~\text{sensitivity} = \dfrac{\text{TP}}{\text{TP}+\text{FN}},\\  
	\end{split}
	\end{equation}
	where TP, TN, FP, and FN are defined the number of true positive, true negative, false positive, and false negative detections, respectively. 
	Moreover, ROC (Receiver Operating Characteristic) curves and their AUCs (Area Under the Curve) is also used to evaluate the performance of classification model. ROC curve is produced by false positive rate (horizontal axis) and true positive rate (vertical axis). A better performance is achieved with a larger AUC.
	
	\subsubsection{Experimental Results}
	\begin{table}[!t]
		\renewcommand{\arraystretch}{1.1}
		\caption{Experimental results on INbreast dataset.}
		\label{table3}
		\centering
	\begin{tabular}{ccccc}
		\toprule
		Method & AUC & Acc. &Sensitivity & Specificity\\
		\midrule	
		\emph{Pred(CNN) }      &0.859  &0.85  &0.828    &0.885 \\ 
		\emph{Pred+Desc}            &0.892  &88.5  &0.863    &\textbf{0.922}\\
		\emph{Pred+Desc+Pres}       &\textbf{0.901}  &\textbf{0.902}  &\textbf{0.906}    &0.896\\
		\bottomrule
	\end{tabular}
	\end{table}
	
	For the experiments, 4-fold cross-validation tests have been carried out, which ensures that the samples are tested equally to prevent any bias error. The real samples are augmented into 512 samples and divided into four folds. Each fold contains 128 real samples with 48 benign masses and 80 malignant masses. Three folds are used for training and the rest for testing. Firstly, we perform predictive binary classification on real data using the CNN model where we term it as \emph{Pred (baseline)}. Then we apply the descriptively trained cGAN to generate synthetic mass images. In addition, we perform predictive learning using CNN on the combination of real and synthetic images for binary classification where we term it as \emph{Pred+Desc}. Finally, to validate the effectiveness of utilizing domain knowledge by prescriptive learning, we apply the introduced DCGAN to generate benign and malignant binary masks followed by synthetic mass image generation from masks through descriptively trained cGAN. As a result, a synthetic dataset with 1040 benign samples and 1636 malignant samples are formed that is combined with three folds of real images to form a new collected set for training the CNN model. We term this procedure as \emph{Pred+Desc+Pres}. All the testings are conducted on the same real data with the same training parameters setting. Experimental results are listed in Table \ref{table3}. 
	
	As shown in Table \ref{table3}, conventional CNN trained on real images with augmentation for  malignancy classification achieves an average accuracy of 0.85. After descriptive adversarial learning for synthetic mass image generation for training, its average accuracy improves to 0.885. In addition, the specific type of binary masks and related mass images are prescriptively generated in a knowledge-driven way to enlarge the variations of training data. In this step, medical knowledge such as that benign mass always arises with oval shape and circumscribed margin, while the malignant mass along with irregular shape and spiculated margins is utilized for data generation and collection. Through such closed-loop optimization on existing real data, an average accuracy of 0.902 is further achieved. The CNN trained on the descriptive and prescriptive generated data performs better with higher classification accuracy than the model trained on the real set under the same testing set and training parameters setting. Besides, as shown in Table \ref{table3}, the AUC improvements also demonstrate the effectiveness of the proposed optimization framework. By further investigation, predictive learning from descriptively generated samples in the artificial imaging system can boost the performance in computational experiments for medical knowledge extraction. Generating the desired binary mask of masses for image synthesis based on the prior medical knowledge through prescriptive learning in our proposed GANs-based PMI framework further improves the accuracy and it can enhance the interpretability for diagnosis. This GANs-based PMI framework can be easily generalized to other tasks of medical imaging. In summary, our proposed data-knowledge-driven PMI framework is capable of describing, predicting, and prescribing the correlation between the image data and medical knowledge in real complex imaging systems. 
	
	\subsection{Analysis of skin lesion images}

\begin{figure*}[!tbp]
	\centering
	\includegraphics[width=0.8\linewidth]{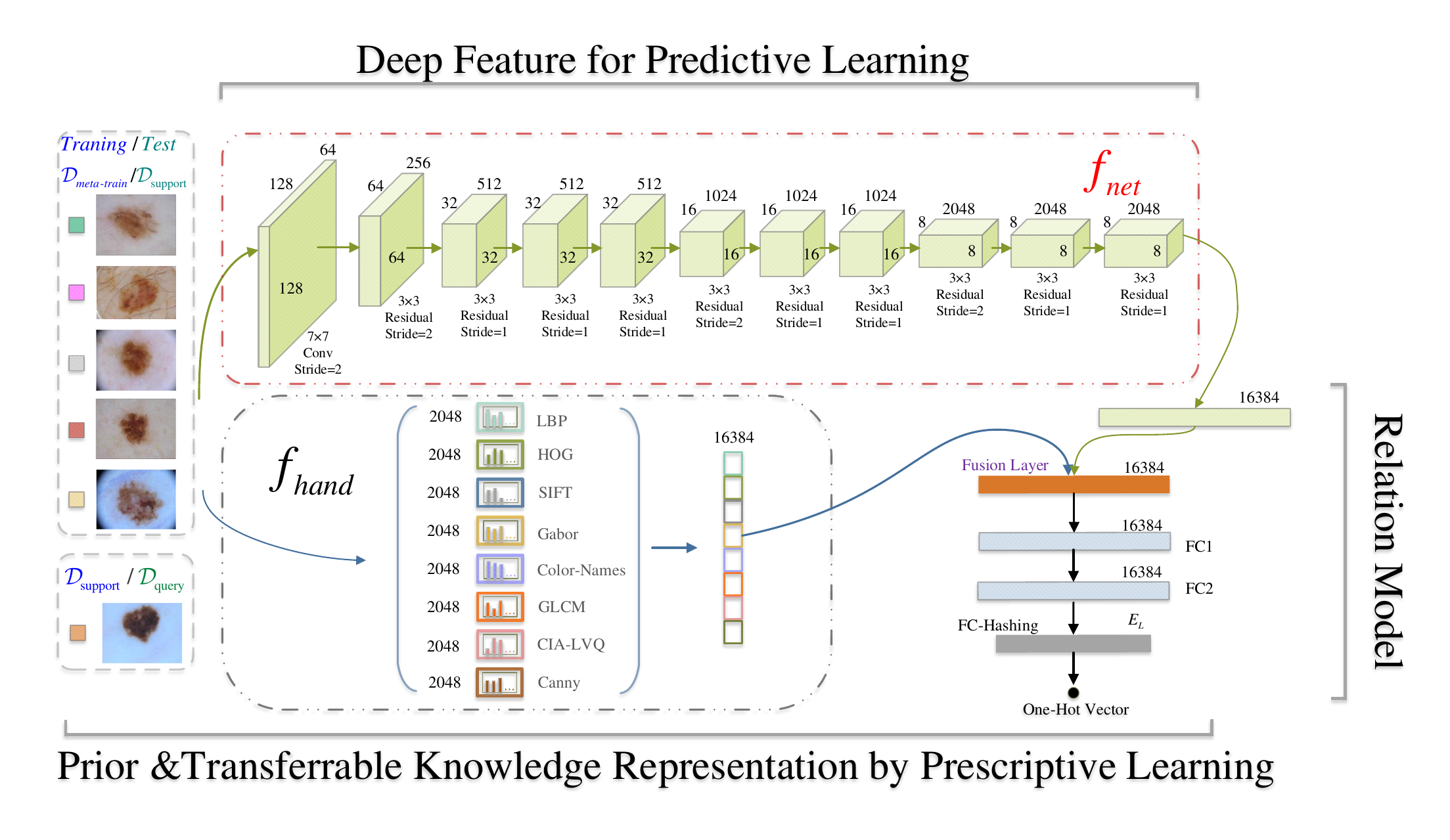}
	\caption{Relation network architecture in PMI for the skin lesion classification.}
	\label{RN_PMI}
\end{figure*}

	\begin{table*}[t]
	\renewcommand{\arraystretch}{1.1}
	\caption{Experimental results on ISIC Skin 2017.}
	\label{table4}
	\centering
	\begin{tabular}{ccccc}
		\toprule
		Methods & AUC & Acc &Sensitivity & Specificity\\
		\midrule	
		\#2      &0.856  &0.824  &0.103    &0.998 \\ 
		\#1      &0.868  &0.828  &0.735    &0.851 \\
		SDL\cite{zhang2019medical}		       &0.868  &0.872  &-         &- \\
		ARLCNN\cite{zhang2019attention}        &0.875  &0.850  &0.658    &0.896 \\
		\midrule	
		\emph{Pred }                     &0.667  &0.732  &-    &- \\ 			
		\emph{Pred+Pres}                 &0.883   &0.890  &0.732    &0.901 \\
		\emph{Pred+Pres+Desc}            &\textbf{0.912}   &\textbf{0.906}  &\textbf{0.743}    &\textbf{0.907} \\
		\bottomrule
	\end{tabular}
\end{table*}

	In last case study, we mainly focus on data-driven for mass classification in mammograms where our major concern is the data generation. For the step of prescriptively learning, we only prescriptively generate specific malignant/benign mask based on the prior knowledge that malignant mass also comes with irregular shape and spiculated margin and benign mass with oval shape and circumscribed margin. In this case of skin image analysis, we try to focus more on knowledge-driven for skin image classification based on PMI. 
	
	In particular, based on PMI framework, we introduce a deep relation model embedded with hand-crafted features \cite{zheng2019relationSkin} for skin lesion classification. The key motivation of employing domain knowledge is in two folds. First, dermatologists classify the skin lesion based on the symptoms of color, margin, texture, shape, and so on. Hence, we propose to extract hand-crafted features to capture these statistical information as prior domain knowledge for diagnosis support. Second, a deep meta-learning strategy is introduced in this work to extract transferrable knowledge.
			
	\subsubsection{Implementation}
	For the step of descriptive learning for synthetic data generation, simple data augmentation of random rotation, horizontal and vertical flips, and scaling are used to enlarge the training dataset to alleviate the problem of over-fitting 
	
	For predictive learning with prescriptive knowledge, inspired by the relation model \cite{sung2018learningRelation} to employ transferrable knowledge for classification,	we introduce a deep hashing relation model to leverage the power of prior experience learned from classifying skin lesion images with limited samples. According to the dermatologists' prior experience and existing medical knowledge, clinical dermoscopic symptoms such as number of clinically significant colors, lesion shapes, sizes, local textures are crucial for skin screenings \cite{barata2017development,saez2018statistical,celebi2019dermoscopy}. For instance, a single lesion with symptoms of variegated tonalities of color, asymmetry in shape, or prominent network is more likely to be melanoma\cite{menzies2001short,haenssle2018man}. Since the hand-crafted features are approximations of the visual symptoms based on the mathematical and statistical formulations designed by the humans without any training images, they can be served to represent the domain prior knowledge in this work. Hence, we propose to embed LBP, HOG, SIFT, Gabor, Color-Names, GLCM, CIA-LVQ and Canny as hand crafted-features to capture the prior knowledge of texture, edge, color, and so on\cite{zheng2019relationSkin,zheng2019relationCaricature}. To further jointly employ the domain knowledge prescriptively for skin cancer screening with very few labeled skin lesion images, few-shot learning phase is applied in this work. During meta learning, it learns to learn a deep distance metric for classification where transferrable knowledge is learned \cite{sung2018learningRelation}. In summary, we introduce a novel deep relation network via few-shot learning embedded with hand-crafted features prescriptively for the skin lesion classification. The framework is illustrated in 
	Fig. \ref{RN_PMI}.	Our goal is to regress the relation score $J_{i,j}$ to the ground truth where the mismatched pair with similarity 0 and matched pair with similarity 1.  The mean square error loss are used to train the model and the objective function is
	\begin{equation}\label{bin_objective_skin}
	L = \sum_{i=1}^{n}\sum_{j=1}^{m}[(J_{i,j}-(y_{i}==y_{j}))^2+\lambda||H-sgn(H)||_{p}^{p}],
	\end{equation}
	where $\lambda||H-sgn(H)||_{p}^{p}$ is a penalty term introduced by \cite{su2018greedy}, $\lambda=0.1\times\frac{1}{N K}$ and $N$ is the number of input size and $K$ is the hashing encoding length. In this work, $H$ is the output of fusion layer.
	
	\subsubsection{Dataset of Skin Lesions and Evaluation Criteria}
	Experiments are conducted on the ISIC Skin 2017 dataset \cite{codella2018skin} where it contains 2000, 150, 600 skin lesion images for training, validation, and testing respectively. Lesion images are paired with a gold standard diagnosis, i.e. melanoma, nevus, and seborrheic keratosis. In this work, we  consider the sub-task of melanoma classification (melanoma vs. others). We collected 1320 additional dermoscopy images from ISIC Archive to enlarge the training dataset\cite{zhang2019attention}. Similar to mass classification, the evaluation criteria of  AUC, accuracy,  sensitivity, and specificity defined in \ref{Criteria} are used as performance metrics. 
	
	\subsubsection{Experimental results}
	
	Comparison experiments are conducted with other work to validate the effectiveness. 
	Firstly, we perform predictive binary melanoma classification using conventional relation model with only deep feature $f_{net}$ where we term it as \emph{Pred}. Then, to utilize the domain knowledge through prescriptive learning, we propose to embed the prior hand-crafted feature $f_{hand}$ with deep feature in the meta-learning scheme. Besides, transferable knowledge representation is achieved through few-shot learning in this step. This procedure is termed as \emph{Pred+Pres}. To validate the effectiveness of the introduced PMI, a simple augmentation of training samples is performed to act as the step of descriptive learning. The whole procedure of PMI is termed as \emph{Pred+Pres+Desc}. 
	
	Experimental results and comparisons with ARLCNN\cite{zhang2019attention}, SDL\cite{zhang2019medical} and top two ranking results in the ISIC 2017 challenge leader-board are listed in Table \ref{table4}. As shown in Table \ref{table4}, the introduced PMI framework can achieve better performance with AUC of 0.912 compared with the state-of-the-arts.  Besides, when we further perform prescriptive meta learning with predictive learning, the AUC of \emph{Pred+Pres}  improves 32.4\%  compared with the result from  baseline of \emph{Pred}. And it can achieve the best performance of AUC with 0.883 compared with other work. Through further investigation, the incorporating of hand-crafted features with domain knowledge and  the meta-learning scheme with prior knowledge by prescriptive learning in PMI is critical for this task of melanoma classification.	Simple augmentation in the step of descriptive learning in PMI can further boost the performance slightly. It further validates that the introduced data-knowledge driven PMI framework is effective for medical image analysis.

	\section{Conclusion}\label{conclusion}
	In this paper, we propose a data-knowledge-driven framework termed as PMI for vision-based intelligent medical image analysis. Artificial imaging systems with descriptive learning allow to collect large scale synthetic and real images for training and evaluating the models in the computational experiments. With a knowledge-to-data in a top-down manner through prescriptive learning, we can select and generate specific image data based on the prior or extracted medical domain knowledge. With a data-to-knowledge in a bottom-up inference through predictive learning, we can extract medical knowledge for clinical diagnostic supporting systems. Through parallel execution, a 'large' scale of medial image data is collected from a 'small' set of real images, followed by 'small' intelligence with interpretable medical knowledge extraction. Experimental results from case studies also demonstrate that the data-knowledge-driven PMI scheme alleviates the limitations of a small quantity of available medical images and enhance the interpretability for final diagnosis and prognosis with more descriptive information. 
	
	Future work will focus on expanding the proposed PMI framework beyond diagnosis decision support in medical imaging. For the foreseeable future, the field of parallel medical imaging has tremendous potential to supplement and verify the work of clinicians, train radiologists to be more skilled, perform the surgical planning, apply intra-operative navigation, give personalized medicine recommendation, and visualize medical images with interpretable masks, particularly in the complex field of imaging analytics with complicated diseases.

	
	\ifCLASSOPTIONcaptionsoff
	\newpage
	\fi

	
	\bibliographystyle{IEEEtran}
	\bibliography{gouc_parallel}

\end{document}